\documentclass[10pt,twocolumn,letterpaper]{article}
\pdfoutput=1
\usepackage[pagenumbers]{cvpr} 
\usepackage{graphicx}
\usepackage{amsmath}
\usepackage{amssymb}
\usepackage{booktabs}
\usepackage{float}
\usepackage{enumitem}
\usepackage{enumerate}

\usepackage[pagebackref,breaklinks,colorlinks]{hyperref}

\usepackage[accsupp]{axessibility}  

\usepackage[capitalize]{cleveref}
\crefname{section}{Sec.}{Secs.}
\Crefname{section}{Section}{Sections}
\Crefname{table}{Table}{Tables}
\crefname{table}{Tab.}{Tabs.}

\def\cvprPaperID{4796} 
\def\confName{CVPR}
\def\confYear{2023}

\begin{document}

\definecolor{myorange}{RGB}{255, 165, 0}
\definecolor{mybrown}{RGB}{155, 78, 20}
\definecolor{mypurple}{RGB}{150, 56, 226}
\definecolor{myblue}{RGB}{10, 50, 230}
\definecolor{myred}{RGB}{230, 10, 10}
\definecolor{mygreen}{RGB}{60, 160, 60}
\definecolor{mydarkgreen}{RGB}{30, 110, 30}
\definecolor{myskyblue}{RGB}{31, 119, 180}
\definecolor{mymagenta}{RGB}{139, 0, 139}
\definecolor{mycyan}{RGB}{0, 100, 100}

\newcommand{\ad}[1]{{\color{mybrown}{[Aniket: #1]}}}
\newcommand{\vishwa}[1]{{\color{mygreen}{[Vishwa: #1]}}}
\newcommand{\emma}[1]{{\color{mypurple}{[Emma: #1]}}}
\newcommand{\florian}[1]{{\color{mymagenta}{[Florian: #1]}}}
\newcommand{\ashok}[1]{{\color{mydarkgreen}{[Ashok: #1]}}}
\newcommand{\aggelos}[1]{{\color{myblue}{[Aggelos: #1]}}}
\newcommand{\ollie}[1]{{\color{mypurple}{[Ollie: #1]}}}
\newcommand{\postrebtl}[1]{{\color{mydarkgreen}{[Post Rebuttal: #1]}}}
\newcommand{\delete}[1]{{\color{myred}{[DELETE: #1]}}}

\newcommand{\todo}[1]{{\color{myorange}{[TODO: #1]}}}
\newcommand{\help}[1]{{\color{myred}{[HELP: #1]}}}

\newcommand{\blue}[1]{{\color{myskyblue}{#1}}}
\newcommand{\orange}[1]{{\color{myorange}{#1}}}
\newcommand{\green}[1]{{\color{mygreen}{#1}}}

\def\epsP{\ensuremath{\epsilon^{\prime}}\xspace}
\def\fsx{{\ensuremath{f_s(x)}}\xspace}

\def\mtos{\ensuremath{\epsilon^{\prime}m^{2}/s}\xspace}


\title{Thermal Spread Functions (TSF): Physics-guided Material Classification}

\author{Aniket Dashpute$^{1,3}$, Vishwanath Saragadam$^3$, Emma Alexander$^2$,\\
	 Florian Willomitzer$^4$, Aggelos Katsaggelos$^1$, Ashok Veeraraghavan$^3$, Oliver Cossairt$^{1,2}$\\
	 {\small $^1$Electrical and Computer Engineering, $^2$Computer Science, Northwestern University,}\\
	 {\small $^3$Electrical and Computer Engineering, Rice University,}\\
	 {\small $^4$Wyant College of Optical Sciences, University of Arizona}
	 }
 
\maketitle

\begin{abstract}
%
Robust and non-destructive material classification is a challenging but crucial first-step in numerous vision applications.
We propose a physics-guided material classification framework that relies on thermal properties of the object.
Our key observation is that the rate of heating and cooling of an object depends on the unique intrinsic properties of the material, namely the emissivity and diffusivity.
We leverage this observation by gently heating the objects in the scene with a low-power laser for a fixed duration and then turning it off, while a thermal camera captures measurements during the heating and cooling process.
We then take this spatial and temporal ``thermal spread function" (TSF) to solve an inverse heat equation using the finite-differences approach, resulting in a spatially varying estimate of diffusivity and emissivity.
These tuples are then used to train a classifier that produces a fine-grained material label at each spatial pixel.
Our approach is extremely simple requiring only a small light source (low power laser) and a thermal camera, and produces robust classification results with $86\%$ accuracy over $16$ classes\footnotemark{}.
\footnotetext{Code: \url{https://github.com/aniketdashpute/TSF}}
\end{abstract}


\section{Introduction}

\begin{figure}
    \centering
    \includegraphics[width=\linewidth]{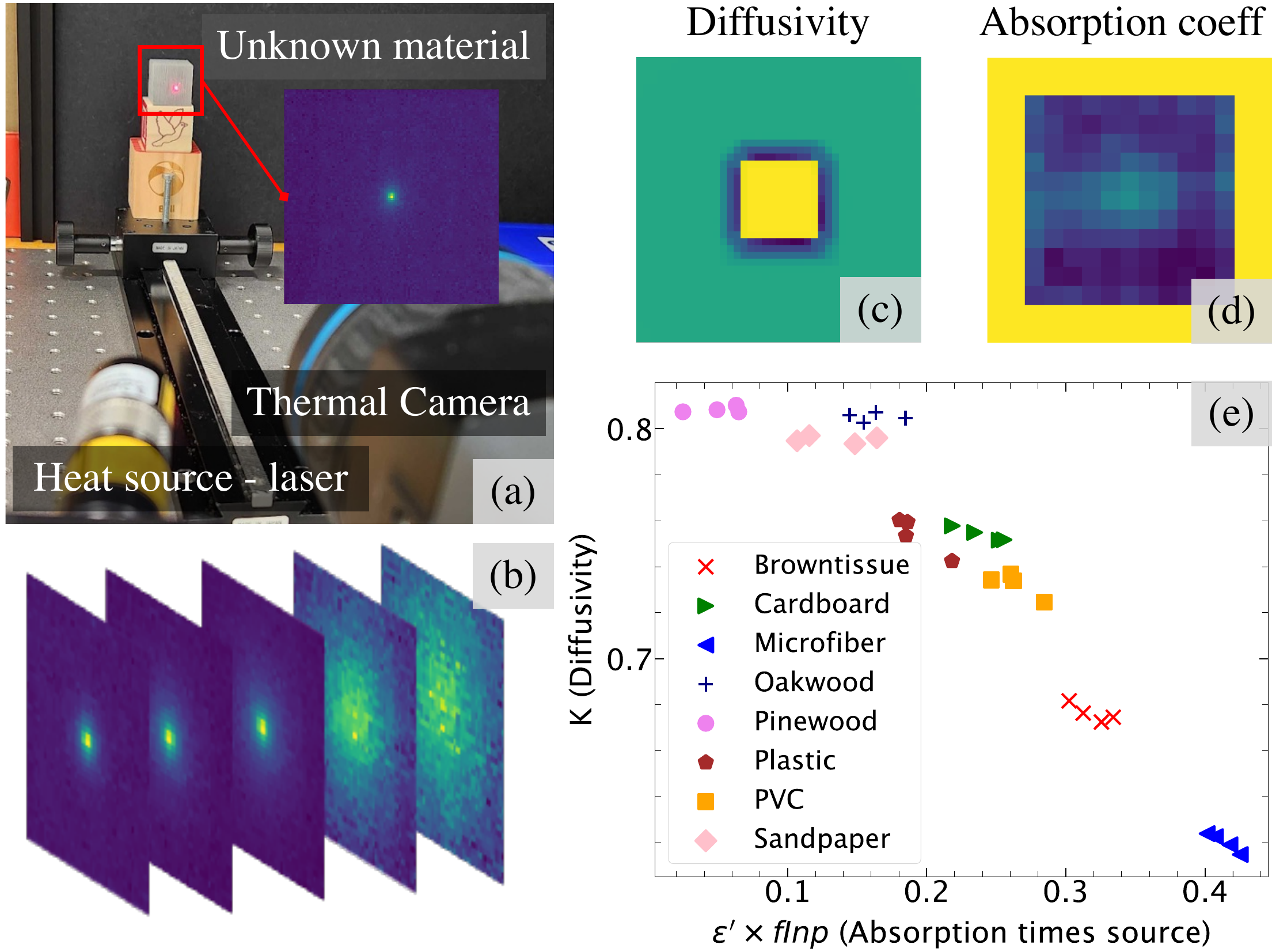}
    \caption{\textbf{Material classification with thermal properties.} Materials have unique thermodynamics that enables robust classification. We propose a simple setup composed of (a) a 60mW laser as a heat source and a thermal camera to capture the heat profile (inset). This results in a (b) stack of images we call  Thermal Spread Function (TSF) that encodes heating and cooling effect around the laser dot. The TSF data is used to (c) estimate diffusivity map and (d) the external heat source term using inverse Finite Difference Method, which is then used to (e) classify the material robustly.}
    \label{fig:Cover}
    \vspace{-5pt}
\end{figure}

Material classification is an important task pertinent to a diverse set of fields including but not limited to medicine and biology~\cite{parida2012classification}, chip manufacturing, recycling~\cite{hendrix1996technologies, buekens2014recycling}, land and weather monitoring using satellites, and vision and robotics.
Robust material classification is particularly critical in separating various parts of an object based on their constituent materials~\cite{hendrix1996technologies,buekens2014recycling}.
Common tools for material classification span a large spectrum including simple tools such as infrared spectroscopy, hyperspectral imaging to more exotic tools such as ultrasound, and x-ray fluorescent imagers.


Material classification primarily relies on various dimensions of light including bidirectional reflectance function (BRDF) slices \cite{wang2009material}, color and NIR images \cite{salamati2009material}, frequency and depth-dependent ToF distortion \cite{tanaka2017material}, spectral imaging methods \cite{ibrahim2010spectral, saragadam2020programmable}, multi-modal methods \cite{erickson2020multimodal}, and thermal imaging \cite{saponaro2015material}. 
Methods based on RGB images are popular due to availability of RGB cameras and large labled datasets, but suffer from lack of robustness. In contrast, spectrum-based imaging based methods enable accurate classification but often require complex optical systems such as hyperspectral cameras, and are sensitive to external illumination conditions.


Human perception of materials is often multi-modal, such as relying on touch and vision to accurately classify material.
This act of touching further involves a thermodynamic exchange that relies on the material composition of the object.
A metallic object results in rapid conduction of heat whereas non-metallic objects like ones made of wood result in slower transfer rate.
%
%
Thus the intrinsic thermal properties provides an insight into the material properties that is often missed or confused by vision alone.
Previous work in contact ways of knowing conductivity - haptic sensing \cite{bednarek2019robotic, zhang2022thermal}, and haptic displays \cite{guiatni2009thermal, singhal2016development} leveraged this idea by developing ``an artificial fingertip". The drawback of this method is that it is invasive - it requires to touch the scene and thus can lead to interfering with it. Thermal characterization for recycling has also been done using a spectrometer and a fluxmeter \cite{tilioua2018characterization}.

Thermal Imaging methods enable contact-free estimation of thermal properties, thus allowing us to classify materials rapidly and in a non-destructive manner. One of the most popular contact-less methods for determining thermal diffusivity is the laser flash method. A laser is flashed on a thin slice (microns thick) of a material and the temperature change is observed from the other side, providing a quantitative estimate of the thermal diffusivity or conductivity \cite{parker1961flash, cowan1963pulse}.
This is restrictive due to the constrained lab setup and requirement of thin slices. Thermal imaging has also been used for non-destructive infrastructure inspection where the difference in thermal behaviour of unaltered and defected zones allow defect detection \cite{garrido2020thermographic}.

We take inspiration from the contact-less methods and develop a non-invasive thermal imaging system for material classification. As opposed to previous methods, our method is robust enough to be used in uncontrolled environments, and not limited to constrained lab setups. We use a visible laser beam as an external heat source that shines on a material, which absorbs a fraction of this beam corresponding to optical wavelength . The absorption of this energy leads to a rise in temperature that shows up in the long wave infrared domain (LWIR) and is captured by the thermal camera. 
%
The thermal camera is used to capture the heating process, and once the heat source is off, its cooling  (refer \cref{fig:Cover}). We define the temperature transients obtained from the heating-cooling as its \textbf{Thermal Spread Function (TSF)} and use it for robustly classifying materials. 

A key challenge with using TSF for classifying materials is that a thermal camera requires a known emissivity $(\varepsilon)$ (ratio of radiated energy of the object to that of a black-body) for accurately estimating the temperature. 
To overcome this ambiguity, we leverage a physically accurate heat diffusion equation (see \cref{sec:HeatDiffusion}) that carefully models the thermodynamic interactions between the ambient scene and the object.
This estimated TSF is then used for training a material classifier which enables robust material classification.



\subsection*{Our approach and main contributions}

When objects are heated through radiation on surface and allowed to cool down, they display characteristic temperature changes. These changes are based on their initial temperature, surface absorption, and heat diffusivity. We inject heat through a small portion on the surface of a material which diffuses throughout the body over time. If we observe a small patch of material in the vicinity of injection, we observe the diffusion - both during the injection phase and during the cooling phase after no external heat is supplied. We call this varying 2D temperature profile as the Thermal Spread Function (TSF) of the material.

We measure the TSF of the material through a Long Wave Infrared (LWIR) thermal camera. We derive diffusivity and an absorption factor from the TSF to characterize the material as these properties are independent of the initial temperature of the object.  Our main contributions are the following.
\begin{itemize}[leftmargin=*]
    \setlength{\itemsep}{1pt}
    \setlength{\parskip}{0pt}
    \setlength{\parsep}{0pt}
    \item We first derive a physically accurate model that characterizes the Thermal Spread Functions (TSFs) as a function of initial temperature of the object and an object's thermodynamic properties.
    \item We then use a Finite Differences (FD) Method to solve the inverse heat problem for recovering parameters related to diffusion, absorption and emission
    \item Finally, we design and demonstrate a simple optical setup for non-invasively recovering the thermodynamic properties and using them to classify materials.
\end{itemize}

\section{Background} \label{sec:Background}

\begin{figure}
    \centering
    \includegraphics[width=8cm]{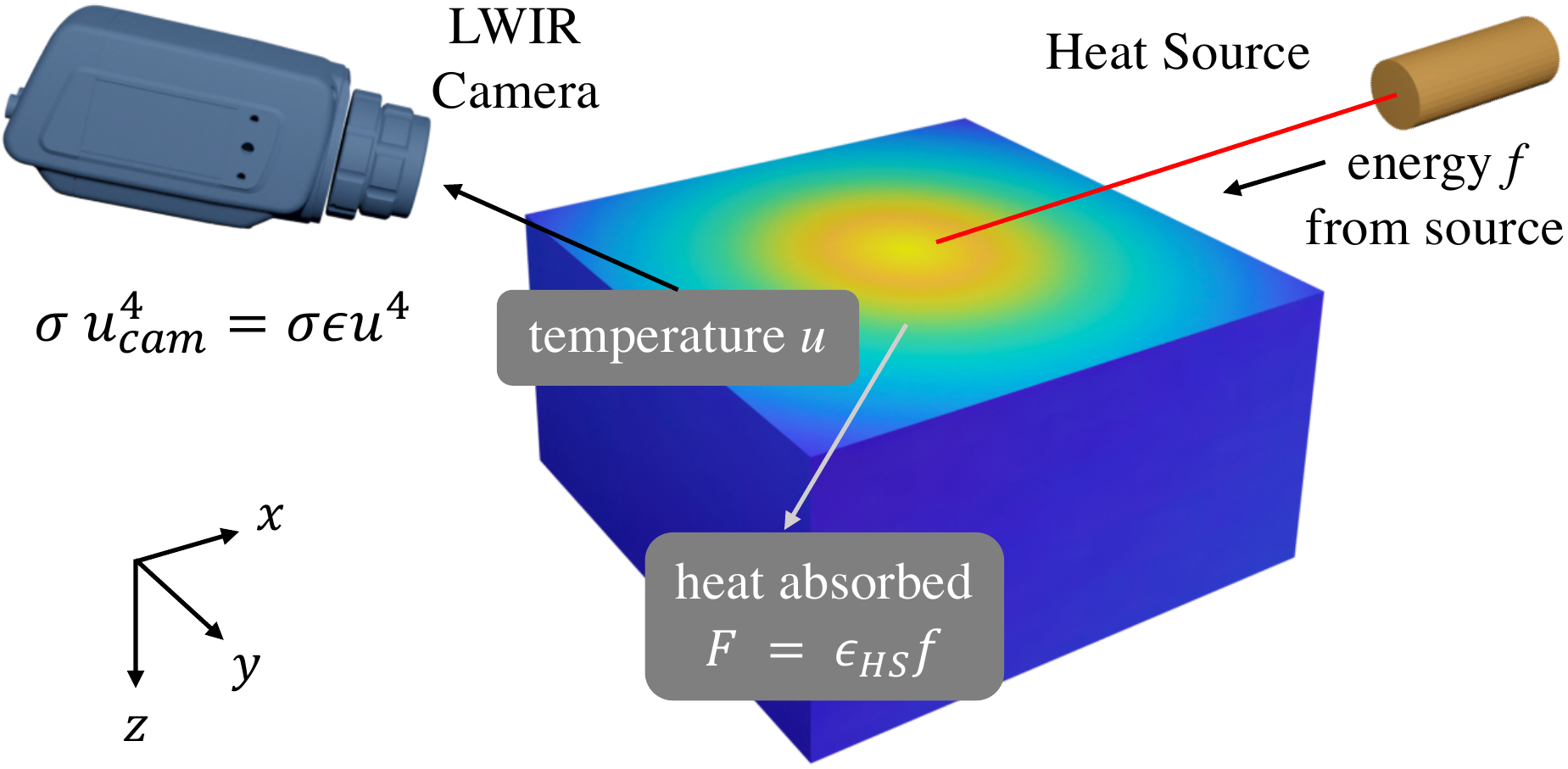}
    \caption{\textbf{Heating and capturing process}. We use an external heat source to heat the object and a thermal camera to observe the heating and cooling effect. Refer \cref{sec:Background} for detailed explanation.}
    \label{fig:Physics}
    \vspace{-5pt}
\end{figure}

We discuss the relevant heat equations and their physical significance below. For more detailed steps and explanation, please refer to \cite{bergman2011introduction, haberman2003applied}.

\subsection*{Heat Diffusion} \label{sec:HeatDiffusion}
Consider the scene shown in \cref{fig:Physics} where an object is illuminated by an external heat source. Let $e$ denote the thermal energy density, $\phi$ represent the heat flux (energy per unit time per unit surface area) across the object boundary and $F$ be the heat absorbed per unit time per unit volume.
The energy conservation equation can be written as,
\begin{equation}
    \frac{\partial e}{\partial t} = - \frac{\partial \phi}{\partial x} + F.
    \label{EqEnergyConservation}
\end{equation}
%
The expression relating thermal energy density and temperature $u(x,t)$ is,
\begin{equation}
    e(x,t) = c(x)\cdot \rho(x)\cdot u(x,t),
    \label{EqHeatEnergy2Temp}
\end{equation}
where $c(x)$ is the specific heat of the material at $x$, and $\rho(x)$ is its density.
Fourier's law of conduction gives a relation between temperature $u(x,t)$ and flux $\phi(x,t)$:
\begin{equation}
    \phi = - \sigma_0\cdot \frac{\partial u}{\partial x},
    \label{EqFourierLaw}
\end{equation}
where $\sigma_0$ represents conductivity of the material.
Substituting \Cref{EqHeatEnergy2Temp,EqFourierLaw} into \Cref{EqEnergyConservation}, we get,
\begin{equation}
    c\cdot \rho\cdot \frac{\partial u}{\partial t} = \frac{\partial}{\partial x}\left(\sigma_0 \frac{\partial u}{\partial x} \right) + F,
    \label{EqHeatEq2Temp}
\end{equation}
where $F$ is the heat received by the material per unit time per unit volume.
Substituting $k=\frac{\sigma_0}{c\rho}$ and $\beta = \frac{1}{c\rho}$ we get,
\begin{equation}
    \frac{\partial u}{\partial t} = \frac{\partial}{\partial x}\left(k\frac{\partial u}{\partial x}\right) + \beta F,
    \label{eqHeatEq1}
\end{equation}
If we inject heat energy $f = f_{external}$ per unit time per unit volume, and $\epsilon_{hs} = \epsilon_{HeatSource}$ is the material absorptivity for the wavelength corresponding to heat source, then $F = \epsilon_{hs} \cdot f$. This denotes that the heat $F$ absorbed by the material is a fraction of the heat $f$ that is supplied.

Our goal is to estimate $k, \varepsilon$ from the observations $u(x, t)$ measured with a thermal camera.
However, a thermal camera can only measure radiance and not the temperature directly.
To solve this, we start by assuming that the absorptivity of the object is $\varepsilon = 1$ which gives us the radiation of an ideal black-body, $u_C=u_\text{black-body}$.
The relationship between the real and ideal black body temperature is $\sigma u^4_\text{black-body} = \sigma \varepsilon u^4$, where $\sigma$ is the Boltzmann constant. This gives us,
%

%
\begin{subequations}
\begin{equation}
    \varepsilon^{-1/4} \frac{\partial u_C}{\partial t} = \frac{\partial}{\partial x}\left(k\frac{\partial}{\partial x}\left(\varepsilon^{-1/4} u_C\right) \right) + \beta\cdot \epsilon_{hs} \cdot f.
    \label{eqHeatEq2}
\end{equation}
We now have an expression relating the measurements of the thermal camera $u_C$, the input light source $f$, and the properties of the material at each spatial location $\varepsilon$ and $k$.
Our goal is to use the measurements to estimate a spatial distribution of the thermal properties, $\varepsilon(x, y)$ and $k(x, y)$.

\Cref{eqHeatEq2} is an underconstrained problem as it has two unknowns for every equation.
To regularize the problem, we assume that the diffusivity and absorptivity are constant over a small neighborhood giving us,
%
%
\begin{equation}
    \frac{\partial u_C}{\partial t} = k\frac{\partial^2 u_C}{\partial x^2} + \frac{\beta \cdot \epsilon_{hs}}{\varepsilon^{-1/4}} f = k\frac{\partial^2 u_C}{\partial x^2} + \epsilon^{\prime}\cdot f,
    \label{eqHeatEq3}
\end{equation}
where $\epsilon^{\prime} = \frac{\beta \cdot \epsilon_{hs}}{\varepsilon^{-1/4}}$.
\end{subequations}
\begin{subequations}
Using the following shorthand notations:
\begin{equation}
    u_t = \frac{\partial u_C}{\partial t}, \text{ }
    u_{xx} = \frac{\partial^2 u_C}{\partial x^2},
    \label{eqU_t}
\end{equation}
\Cref{eqHeatEq3} can be written as:
\begin{equation}
    u_t = k \cdot u_{xx} + \epsP \cdot f
    \label{eqHeatEqnFinal1D}
\end{equation}
\end{subequations}

We derived all the equations above for the one-dimensional case for simplicity, but the analysis can be easily extended to 3 dimensions. By replacing the Laplacian term $u_{xx}$ from 1-D \Cref{eqHeatEqnFinal1D} by the 3-D Laplacian $\Delta u_{xyz} = u_{xx} + u_{yy} + u_{zz}$, we obtain:

\begin{equation}
    u_t = k \cdot \Delta u_{xyz} + \epsP \cdot f
    \label{eqHeatEqnFinal}
\end{equation}

\Cref{eqHeatEqnFinal} is central to material classification proposed in this paper. The term $k$ denotes thermal diffusivity of a material and \epsP is a factor that depends on absorption (in heat source domain) and emission (in LWIR domain). 
The tuple of parameters forms a unique signature for materials that we use for the downstream task of classification.

\subsection{Modeling the source} \label{sec:ForwardModel}
We use an external source that is switched on for a known $t_{ON}$ duration and switched off. Let $\textbf{x} = (x, y, z)$ be the combined spatial variable. We model $f(\textbf{x},t)$ as:

\begin{equation}
f(\textbf{x},t) = 
\left\{
    \begin{array}{lr}
        f_s(\textbf{x}), & \text{for } t \le t_{ON}\\
        0, & \text{for } t > t_{ON}
    \end{array}
\right\}
\end{equation}
where $f_s(\textbf{x})$ is the spatial profile of the external heat source.
%

For simplicity, we make the following assumptions:
\begin{itemize}[leftmargin=*]
    \setlength{\itemsep}{1pt}
    \setlength{\parskip}{0pt}
    \setlength{\parsep}{0pt}
    \item Our object shape is known and has a flat surface. This constraint can be relaxed by using a method like Structured Light to get the shape of object
    \item The flat surface of the object faces the camera such that $(x,y)$ coordinates lie in the plane of the surface and $z$ lies perpendicular to it with $z=0$ being at the surface and increasing as we go inside
    \item For a given $(x,y)$ the initial temperature is constant for all $z$, that is,
    \begin{equation}
    u(x,y,z,t=0) = u(x,y,t=0) \text{ , } \forall z \ge 0
\end{equation}
\end{itemize}

\subsection{Choosing spatial profile of source $f_s(\textbf{x})$} \label{sec:SourceProfile}

A spatially-uniform, known-intensity heat source allows us to estimate \epsP from the temporal profile of heating and cooling. Estimation of $k$, however, requires spatial variation in temperature so that diffusion can occur. This can be seen from the Laplacian term, which will be zero across a patch of uniform temperature. This problem, in the absence of source, is the diffusion equation, and hence diffusivity recovery from a uniform-temperature patch is mathematically identical to depth from Gaussian defocus on a texture-free patch \cite{guo2017focal, alexander2016focal}. Therefore, a useful illumination pattern must contain spatial gradients to enable robust estimation of diffusivity parameter. So we decided to use focused points on the scene that generate the required heat profile and hence the texture to get gradients.



\subsection{Inverse Heat Problem} \label{sec:InvProblem}

Given $k$, \epsP, $f_s(x)$ and initial conditions, we can leverage the forward model to obtain temperature profile throughout the volume.
However, we can only capture the surface measurements at various time steps. This requires us to solve the inverse heat problem of recovering the thermal parameters $k$ and \epsP.
This is a challenging underconstrained problem. We relax the problem to make it invertible by leveraging assumptions made in \cref{sec:ForwardModel}.
%
%
%
Given the assumptions, we can use Finite Difference method~\cite{morton2005numerical}, Green's function-based approach~\cite{haberman2003applied}, or a simple curve fitting-based approach to solve for the spatial distribution of $k,\epsP$. We discuss their details in \cref{sec:FiniteDiff}.

\section{Proposed approach}


\subsection{Heating \& Cooling - Thermal Spread Functions} \label{sec:TSF}

\begin{figure}
    \centering
    \includegraphics[width=\linewidth]{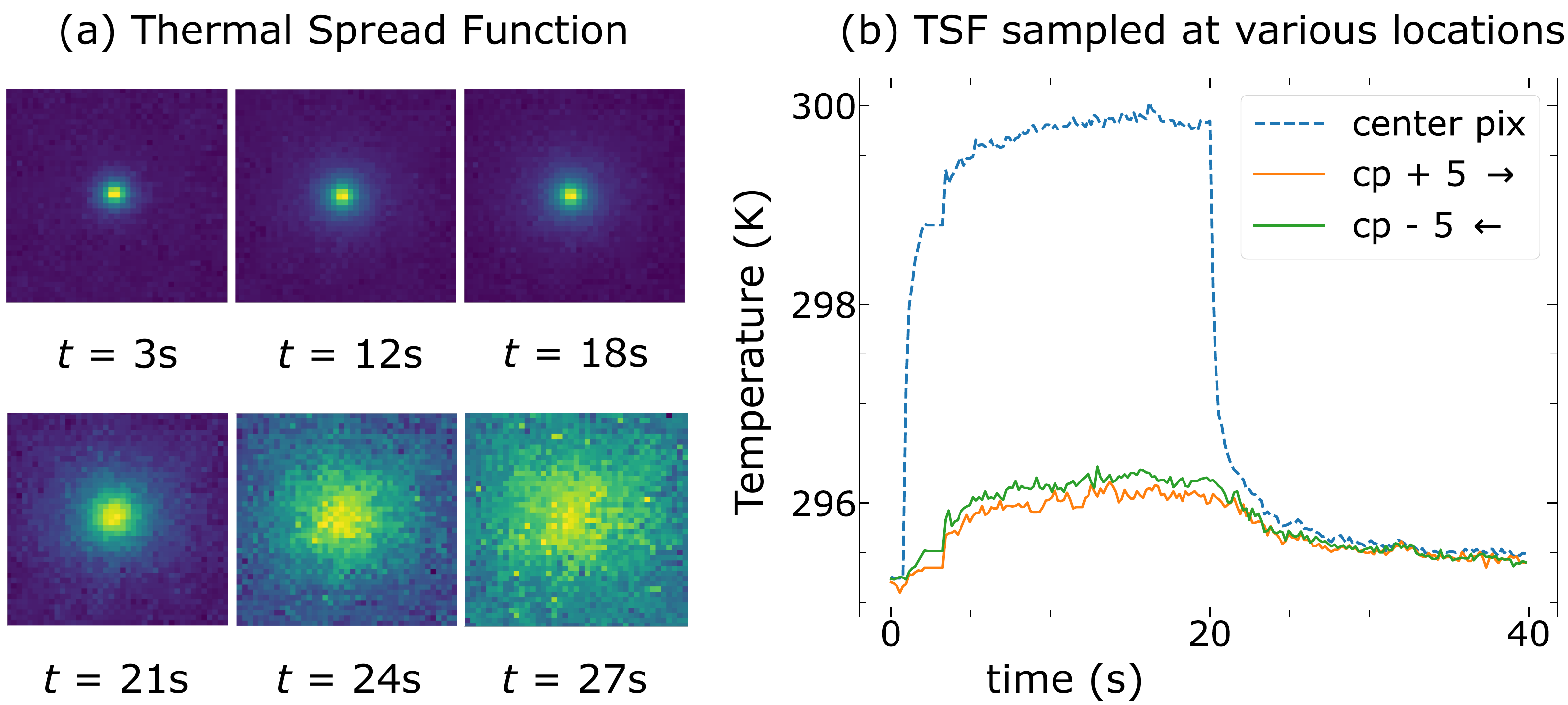}
    \caption{\textbf{Thermal Spread Function.} (a) The images captured by the thermal camera as the laser was switched on (top row), and then off (second row) are shown. We call this stack of images the thermal spread function as it encodes the spatial and temporal spread of thermal radiation in the object. In (b), \blue{dashed blue} line graph on top shows the measured temperature at the point where the laser was projected and the bottom \green{solid green} and \orange{solid orange} line plots show temperature 5 pixels away from center. TSF carries a unique signature about the material which is evident from the temperature plots.}
    \label{fig:TSF}
    \vspace{-5pt}
\end{figure}

\begin{figure*}
    \centering
    \includegraphics[width=17cm]{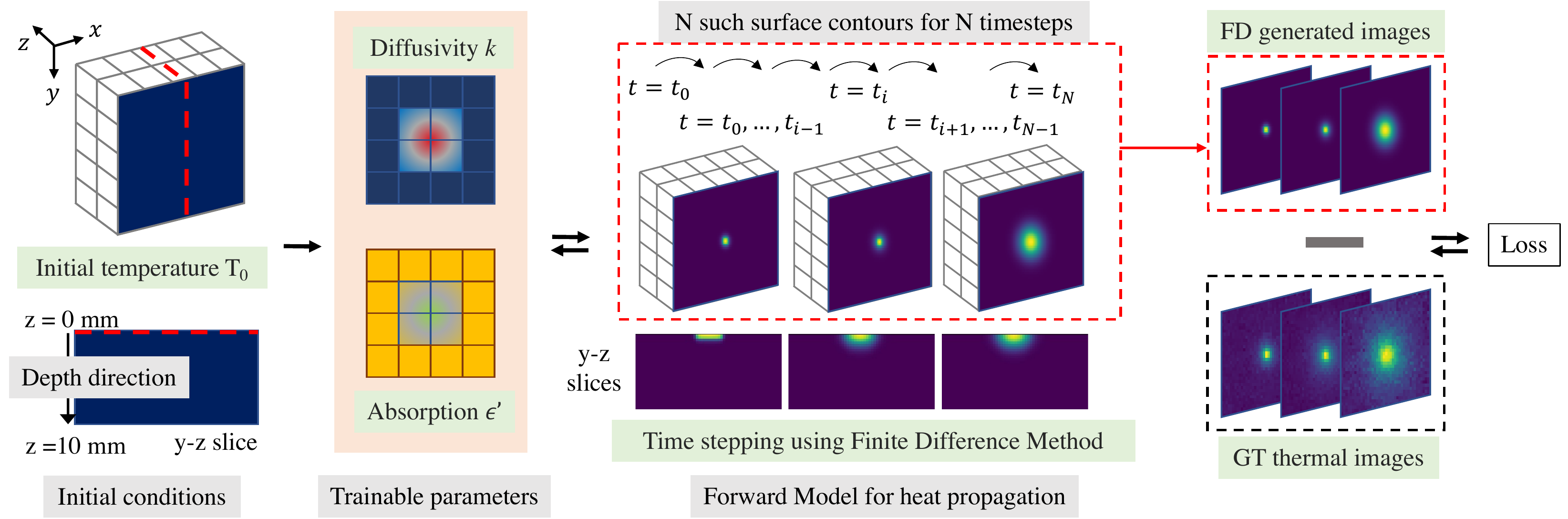}
    \caption{\textbf{Differentiable finite difference (FD) method for estimating thermal parameters.} Computing the diffusivity and emissivity of an object requires solving the heat equation. We propose a novel iterative approach using FD method. We implement a forward operator that takes diffusivity and absorptivity as inputs and uses FD method to generate the TSF for a given set of time stamps. The generated TSF is compared against the measured TSF from the camera with an MSE loss function. We then perform multiple iterations to refine the estimates of diffusivity and absorptivity. Since we leverage physics of TSF, the results are highly accurate, and robust to noise in measurements.}
    \label{fig:FDMethod}
    \vspace{-8pt}
\end{figure*}

A known external heat source $f_s(x)$ that injects heat into the materials for time $t_{ON}$ increases the temperature of the object.
The injected heat will increase temperature at the points of injection and this rise in temperature leads to diffusion of heat across the rest of the material.

This combined phenomenon of absorption and diffusion creates a distinct temperature profile over the surface that is observed by the thermal camera.
This temperature profile varies over time as more heat is absorbed, leading to increase in the temperature variation over the material, which in turn leads to more diffusion.
Once the heat source is switched off, heat diffusion occurs until all points in that object are at the same temperature. 
A thermal camera measures this time-varying temperature on the surface of the object which we define as the Thermal Spread Function (TSF) of the material. For a given material, the TSF depends on initial temperature, and the spatial profile and intensity of external heat source $f_s(x)$. \cref{fig:TSF} demonstrates the TSF for some common material in a scene.

We parameterize the TSF as a function of the parameters $k$ and \epsP from \cref{eqHeatEqnFinal}. This gives us two-fold advantage, (1) it spares us from processing large data for each TSF directly for the classification problem, and (2) TSFs are dependent on initial temperature, so TSF by themselves cannot be used without accounting for the change in initial temperature.

\subsection{Finite Difference Method for Inverse Heat problems} \label{sec:FD} \label{sec:FiniteDiff}

Given initial conditions, k, \epsP, and external heat source \fsx, we can use many heat propagation methods to obtain the temperature distribution over time. Some of the methods include Finite Element Analysis (including FEA Softwares like Ansys, Abaqus), convolving with heat kernels (Green's functions) or Finite Difference (FD) Numerical Methods.

We choose FD method for our forward propagation model as they enable a computationally tractable way to obtain k and \epsP via gradient descent-based optimization. We present below a short analysis of the steps for forward propagation using FD (please refer \cite{haberman2003applied, morton2005numerical} for detailed analysis).
Using the explicit Forward Time, Centered Space method, we obtain time derivatives with a forward difference as:

\begin{equation}
    \frac{\partial u}{\partial t} =  u_t = \frac{u^{t+1}_{x,y,z} - u^{t}_{x,y,z}}{\Delta t},
    \label{EqFDut}
\end{equation}

Similarly, central difference approximation to the Laplacian term can be written as follows (writing only the $x$ term, other terms can be written similarly):

\begin{equation}
    \frac{\partial^2 u}{\partial x^2} =  u_{xx} = \frac{u^{t}_{x+1,y,z} - 2 u^{t}_{x,y,z} + u^{t}_{x-1,y,z}}{(\Delta x)^2},
    \label{EqFDuxx}
\end{equation}

The complete Laplacian term can then be written as a sum of $u_{xx}$, $u_{yy}$ and $u_{zz}$. Substituting \cref{EqFDut,EqFDuxx} into \cref{eqHeatEqnFinal}, we obtain:

\begin{equation}
    u^{t+1}_{x,y,z} = u^{t}_{z,y,z} + \Delta t \left [ k(x,y,z) \cdot \Delta u^t_{x,y,z}  +\epsP(x,y,z) \cdot f^t(x) \right]
    \label{EqFDTimeStepping}
\end{equation}

\cref{EqFDTimeStepping} is the time stepping equation where we obtain the temperature values for time-step $t+1$ given the values at time-step $t$. We start these calculations by applying the initial conditions which give us values at $t=0$. Other relevant boundary conditions are applied during each step as required. Using this method, we can obtain temperature values at any spatial and temporal point by time-stepping. Refer to \cref{fig:FDMethod} for a visualization of the process.


Thus, given the initial conditions, boundary conditions, and all thermal properties, we can define the forward model using FD as follows:

\begin{equation}
    \mathcal{F} : u^{t=0}_{\textbf{x}}, u^{t}_{\textbf{x}=0}, k_\textbf{x}, \epsP_\textbf{x} \longrightarrow u^{t}_{\textbf{x}},
\end{equation}
implying the temperature at $(x,y,z,t)$ can be obtained as $u(x,y,z,t) = \mathcal{F}(k, \epsP, x, y, z, t)$.

This function $\mathcal{F}$ forms the basis of our optimization process. This constitutes our forward model where we optimize for $k$ and \epsP given all the boundary and initial conditions, and external heat source. We constrain this optimization process using the ground truth 2D images we have of the surface temperature of the object over time.

\subsection{Drawbacks of Image Based Analysis} \label{sec:InvHeatImage}

\begin{figure}
    \centering
    \includegraphics[width=\linewidth]{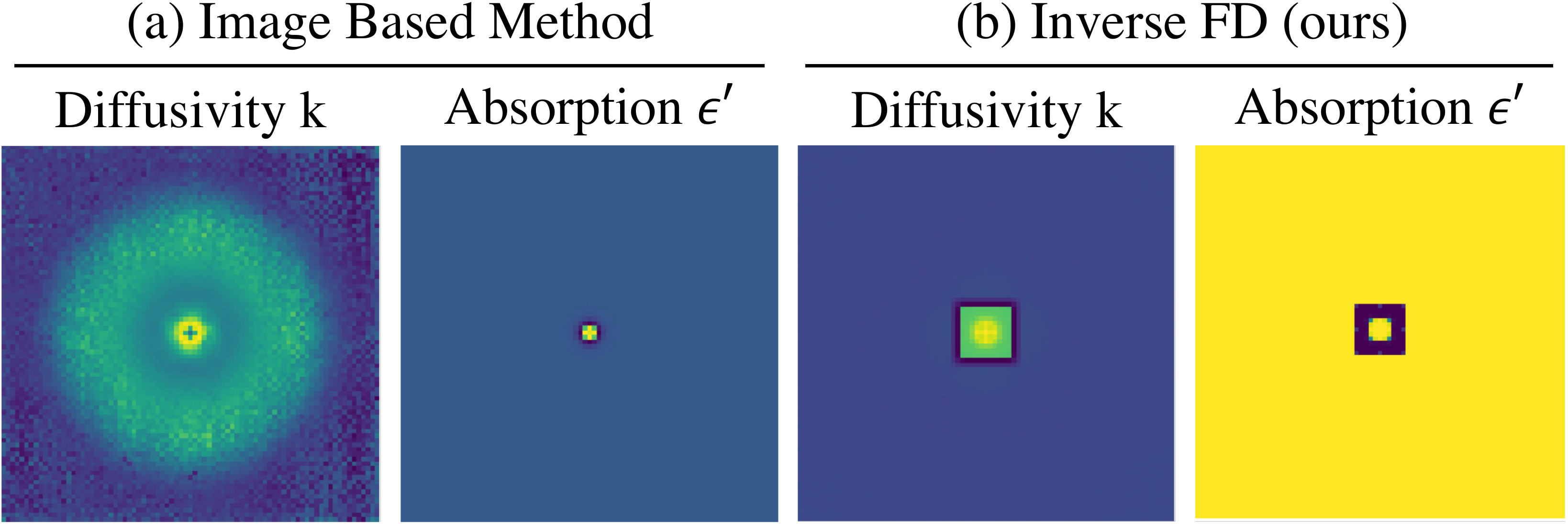}
    \caption{\textbf{Drawbacks of using $u_t$ and $u_{xx}$ directly from images}. Image based method (a) gives a 2D Laplacian term (\cref{eqHeatEqnFinal}) assuming 2D diffusion. This leads to inaccurate donut-shaped results. Our method - inverse FD (b) recovers accurate diffusivity and absorption in the region of heat diffusion by also accounting for diffusion in depth.}
    \label{fig:2DAnalysis}
    \vspace{-5pt}
\end{figure}


As discussed in \cref{sec:InvProblem}, one of the other methods to solve for k and \epsP is to do data-based curve fitting. So we need to solve for k and \epsP given a set of linear equations, where we would know the values of $u_t$ and the Laplacian term $\Delta u_{x,y,z}$. Ideally once we have $u_C(x,y,z,t)$, the time derivative and Laplacian can be calculated using finite difference approximation. The drawback here is that we do not have $u_C(x,y,z,t)$ for all values of $z$ but only for $z=0$. This means from the image based values $u_C(x,y,z=0,t)$, we can get $u_t$ and only the $x$ and $y$ terms of the Laplacian $\Delta u_{x,y,z}$ but not the z term $u_{zz}$. So if we do the data-based curve fitting on this surface data, we will be ignoring the $u_{zz}$ term which leads to inaccurate results. \cref{fig:2DAnalysis}(a, b) demonstrates the inaccuracies observed in the recovered values of diffusivity $k$ over the surface of the material.

Notice the doughnut shaped diffusivity result obtained for $k$ in \cref{fig:2DAnalysis}(a). Intuitively, this is because we ignore the diffusion taking place in the depth dimension.  This is one of the main reasons we shift from the data-based solution to Finite Difference (FD) Method described in \cref{sec:FD}.



\subsection{Classification of Materials based on $k$, \epsP} \label{sec:Classification}

\begin{figure*}
    \centering
    \includegraphics[width=\linewidth]{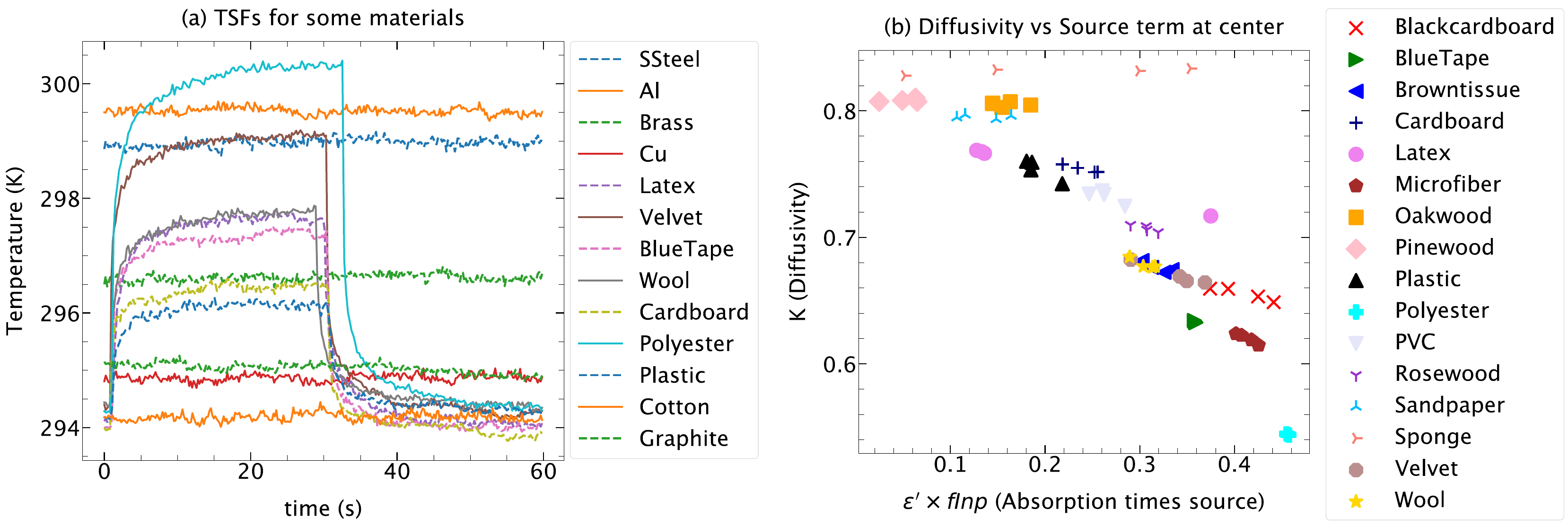}
    \caption{\textbf{Material classification with thermal properties $(k, \epsP)$.} 
    (a) TSFs at center pixel for 13 materials. The curves for heat conductors - Stainless Steel, Aluminium, Brass, Copper and Graphite are almost flat, owing to their high diffusivity and low absorptivity; TSF for cotton is flat due to its high transmissivity. We use the other TSFs to obtain the $(k, \epsP)$ tuples as shown in (b).
    The plot (b) shows clustering of diffusivity and absorptivity for various materials. Expectedly, the tuple of thermal properties form separate natural clusters corresponding to each material which enables us to use simple classification approaches. Here, we use a random forest classifier that takes a few values of $(k, \epsP)$ around the center as input and outputs the material class.}
    \label{fig:Classification}
    \vspace{-5pt}
\end{figure*}



Once we convert our TSF into the thermal parameters, we use those to characterize the materials and classify them into different categories and also identify them. We make 4 categories of materials - fabric, wood, paper, and metals. We have a dictionary of 22 different materials, each material falling into one of the four categories. 

For known materials, we create a scatter plot using the recovered values of diffusivity $k$ and absorption parameter \epsP as shown in \cref{fig:Classification}(b). For each material, we take multiple readings by changing its orientation and initial temperature. Our method is robust to these variations and gives clusters of data points for each of the materials. Once we have this dictionary, we can use the recovered $k$ and \epsP obtained from an unknown material and classify it into one of the material classes.
We use a multi-layer perceptron (MLP, one hidden layer with 90 neurons) as our classifier for its simplicity and speed, and its suitability for small dataset. We train the MLP classifier on the dictionary we create and use that to classify a new material based on $k$ and \epsP.
We also experiment and compare performance with other suitable classifiers - Support Vector Classifiers (SVC, nuSVC), and Random Forest classifier (RF) (refer \cref{tab:ClassifierComparison}, \cref{sec:results}).

\section{Experimental Setup \& Results}


\subsection{Simulations} \label{sec:ExpSim}
 
We use ANSYS Fluent FEM Analysis Software for simulating the heat diffusion process. We use a custom function to define the heat injection which we allow to penetrate to a certain depth. We model this injection to be equivalent of using a relatively high power laser to heat the objects. Modeling the laser means that we also need to simulate the Gaussian beam profile of the heating source.


We can choose one of the existing materials or can create a new one based on the properties from \cref{eqHeatEq1} and \cref{eqHeatEq3}. We model the custom heat injection based on our laser source and choose its intensity accordingly. We assume same surface penetration depth for all materials (the penetration is so low that we can assume it only hits the surface layer of our discretized voxel grid). We run the simulations for 40 seconds, with the heat source being on for $t_{ON}$ = 20 seconds. We use a timestep $\Delta t$ of 0.25s. The total size of our object is 50mm $\times$ 50mm $\times$ 30mm where each discretized voxel size is $\Delta x$ = $\Delta y$ = $\Delta z$ = 0.5mm.

For the simulation process, we define a mesh on a cuboid and only take the surface readings from the simulation. A thermal camera would also provide us only surface readings, so we use these simulated surface contours as ground truth to train our FD model and test its performance. We used a material from ANSYS library (wood) with thermal properties as - density $\rho = 700$ kg/$\text{m}^3$, specific heat $C_p = 2310$ J/(kg K) and thermal conductivity $\sigma = 0.173$ W/(m K). From \cref{eqHeatEq1}, the thermal diffusivity of this material is $1.069 \times 10^{-7}$ $\text{m}^2$/s. When we solve the inverse problem stated above using our FD optimization framework, we obtain a diffusivity value of $1.066 \times 10^{-7}$ $\text{m}^2$/s, which is very close to that of the ground truth value.


\subsection{Experimental Setup} \label{sec:ExpLab}
%
\cref{fig:Cover}(a) shows our experimental setup. We use a FLIR A655sc thermal camera, which has a resolution of 640 $\times$ 480, and works in the LWIR range from 7-15 $\mu$m, and 
a 633 nm, 60 mW laser with an Aruduino controlled relay switch. 
This power rating ensures that the laser is powerful enough to heat while safe enough to use.
We control the system remotely as any warm body present near the setup can hamper with the readings because of thermal radiation.

We use a spatially varying source, a gaussian profiled laser beam. We use a total capture time of 60 seconds, in which the laser in on for $t_{ON}$ = 20 seconds. In the streaming mode of the camera, we capture 300 frames for 60 seconds, implying each time step measures $\Delta t$ = 0.20 seconds. We found that our approach resulted in similar results for a total capture duration of 20s or higher, detailed analysis on this is presented in the supplementary. Our algorithm can also handle multiple simultaneous scans as shown in \cref{fig:2Layers}(a)
%
%
We placed a graded ruler in the scene next to the target objects to calibrate the physical size of each pixel. We perform this calibration every time we move the setup, but on average our value of spatial $\Delta \textbf{x}$ was 0.5 mm (same for all dimensions). This process could be automated using structured light methods (refer \cref{sec:FutureWork}).
%
%

We code the differentiable FD algorithm in python using PyTorch \cite{Paszke_PyTorch_An_Imperative_2019}. We use an ADAM classifier with a decreasing learning rate for 400 epochs. Our system consisting of an RTX 3060, takes 5 minutes for the process to complete and recover the $k$ and \epsP maps. We captured measurements for each material in different initial temperatures and orientations. The code along with the dataset has been released.


\begin{table}[!tt]
\center
\begin{tabular}{lcccc}
\toprule
      \multicolumn{1}{l}{Features} & SVC & nuSVC & RF & MLP\\ \midrule

{ 2} & 68.8\% & 76.6\% & \textbf{82.8\%} & 76.6\%  \\ 
18 & 65.6\% & 82.8\% & \textbf{84.4\%} & 81.2\%\\
50 & 64.0\% & \textbf{85.9\%} & 84.4\% & \textbf{85.9\%}\\
\bottomrule
\end{tabular}
\caption{\textbf{Comparing accuracies of classifiers.} We experimented with various classifiers and number of features for our dataset. We observed that an MLP and nuSVC perform best with 25 features taken each from the recovered $k$ and $\epsP$ images (window size of 5). The accuracy did not improve upon adding further features.}
\label{tab:ClassifierComparison}
\vspace{-10pt}
\end{table}

\subsection{Results} \label{sec:results}

In this section, we analyze the experimental observations made in lab and look at the performance of our parameter recovery and classification algorithms.

Similar to \cite{saponaro2015material}, we chose a few materials from the coarse categories of fabric, wood, paper, and metals. We made a dataset consisting of 22 materials or subclasses, applied our recovery algorithm on the data and used the generated features to train the classifier.
Since metals have very low absorption and high diffusion, their TSF variation tends to have a very low magnitude. Since the \epsP values are almost zero for metals, the FD algorithm did not converge for them. Although we cannot differentiate between different metals, this can be used as an indicator for metals. The data in \cref{fig:Classification}(a) for metals validates the theory for metals. Barring the conducting materials, we performed FD analysis on the other materials and present our analysis on them.

\begin{figure*}
    \centering
    \includegraphics[width=\linewidth]{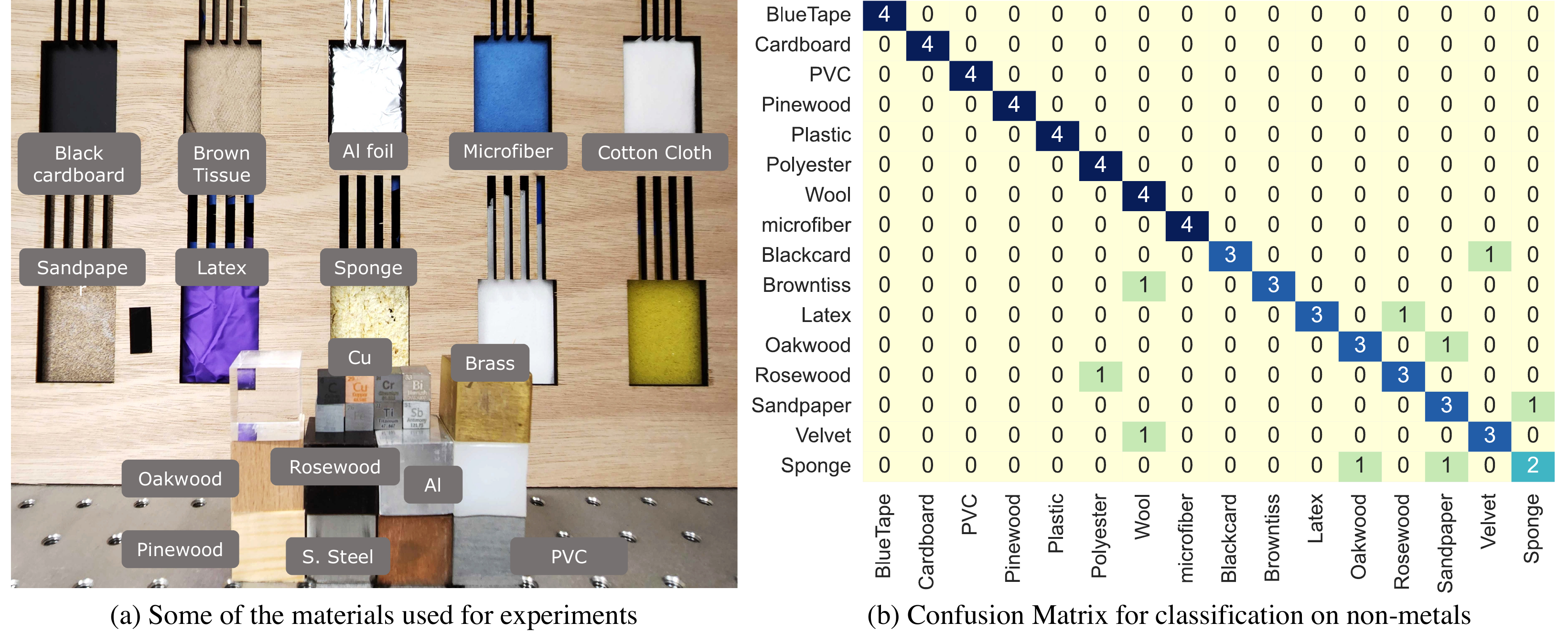}
    \caption{\textbf{Material library and classification results}: We applied our recovery and classification algorithm on our dictionary of materials (a), and show the classification results using confusion matrix in (b). We use random forest as our classifier and use leave-one-out cross validation technique for testing our algorithm. We get an overall accuracy of 81.3\%.}
    \label{fig:matrix}
    \vspace{-5pt}
\end{figure*}

 \cref{fig:Classification} shows a plot of $k$ vs $\epsP$ sampled at the center pixels of their recovered maps. For each measurement, we generate a feature vector by sampling $k$ vs $\epsP$ in a window of specified size around the center. This feature vector along with the material label forms our training set for the classification algorithm. We perform a leave-one-out cross validation approach to obtain the classification accuracy of our method. The classification results can be found in the confusion matrix in \cref{fig:matrix}(b). We obtain an overall accuracy of 85.9\% using a multi-layer perceptron (MLP) classifier with a feature size of 50 pixels. 
 With more features, the subclasses become more separable compared to what we see in \cref{fig:Classification}(b).
 We also compare performance with varying feature sizes and classifiers suitable for dataset \cref{tab:ClassifierComparison}.
 
 Materials that have similar diffusivity, and similar absorptivity for red wavelength (used in our experiments), produce similar TSFs.
 This in turn makes the recovered properties of these materials similar to each other. Such an issue occurs for brown tissue and wool as seen in \cref{fig:Classification}(b) and \cref{fig:matrix}(b) where the algorithm gets confused between the two.
 This can be mitigated by using more lasers like red, green and blue which gives us a set of absorption coefficients enabling more efficient segregation.


\section{Limitations}


Our current modeling of thermodynamics requires certain simplifying assumptions for accurate results, such as planar geometry and constant temperature on the surface and bulk of the objects.
These assumptions have enabled robust classification results in our preliminary experiments, but can be relaxed to extend to complex shaped objects with varying temperature profiles.
Our approach can be combined with approaches for estimating scene geometry such as structured light~\cite{landmann20183d,landmann20203d,erdozain20203d} to estimate material properties of complex scenes and will be pursued as promising future direction.

%

\section{Future Work} \label{sec:FutureWork}

Our experiments covered a limited set of materials to show that their TSFs can be used for material classification. Collecting a larger dataset that covers varying surface color, roughness, and geometry will enable a more robust material classification and will be pursued as future direction.
We used Finite Differences in the forward model of our analysis. This part can also be potentially replaced by other approaches such as Physics Informed Neural Networks (PINNs) \cite{raissi2019physics, cai2021physics, he2021physics}. We can also potentially perform joint estimation of shape and the material properties using thermal structured Light. A known pattern of laser dots on the scene can be leveraged to estimate the thermal properties using the heat diffusion as well as the shape using the texture that is generated.

\subsection{Going Below the Surface} \label{sec:2layers}

\begin{figure}
    \centering
    \includegraphics[width=\linewidth]{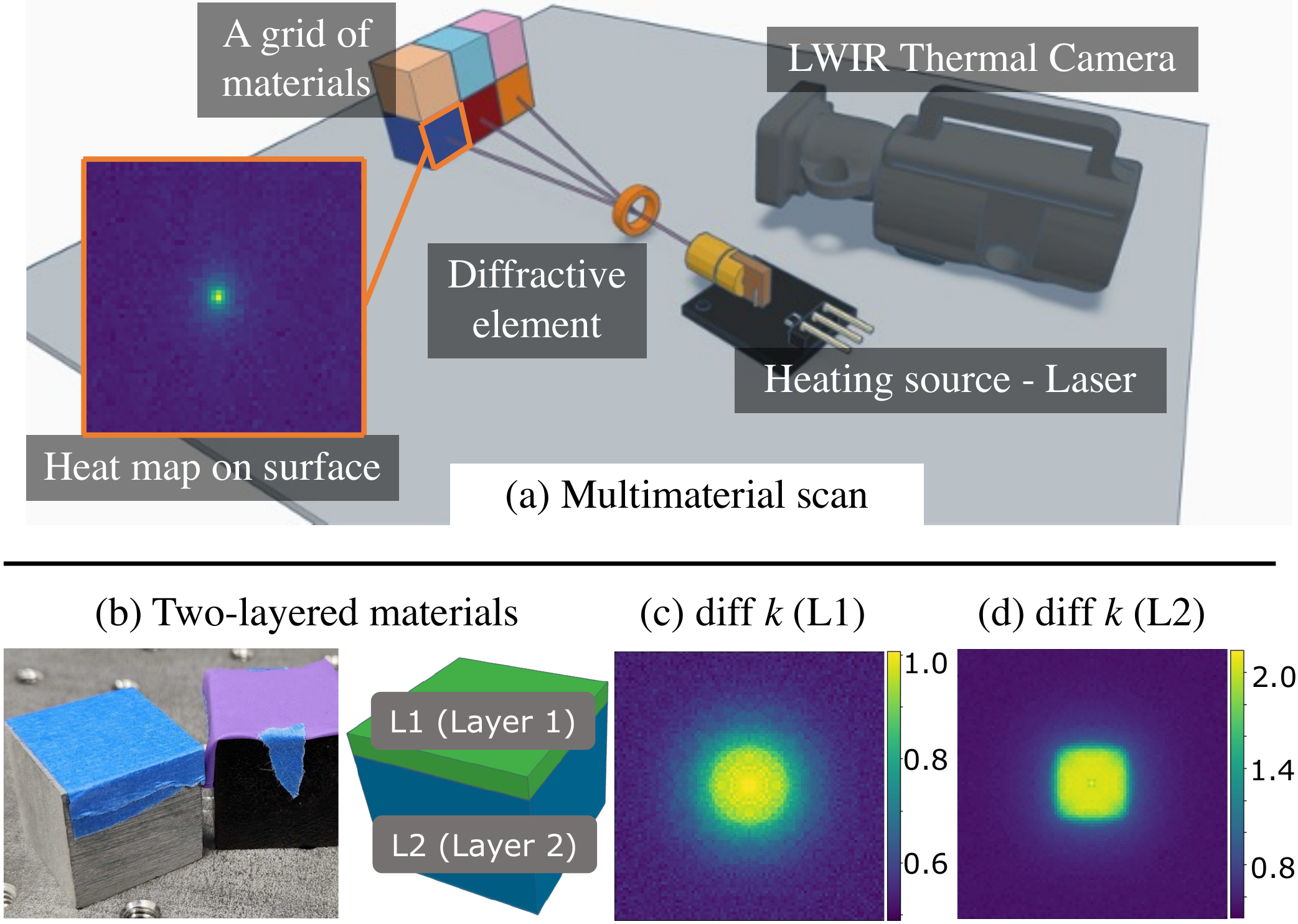}
    \caption{\textbf{Extending our approach.} (a) Our setup can be extended to accommodate simultaneous scans. (b) We ran simulations to understand the performance of our approach on multi-layered materials (b) and to analyse the possibility of recovering thermal properties below the surface. We recovered diffusivity of layers 1 and 2 (c, d) with 0.01\% error, showing promise in "seeing the hidden"}
    \label{fig:2Layers}
    \vspace{-10pt}
\end{figure}

An advantage of thermal material classification over other methods like spectroscopy is that analyzing the heat diffusion can give us information that is hidden to the eye (even the thermal one).
Consider a system as shown in \cref{fig:2Layers}(b), where the top layer is a different material from the one below it. When we inject heat, the upper layer absorbs it based on its surface and absorption properties, while the diffusion is affected by the diffusivity of both the upper layer and the layer underneath.
We analyzed this problem in simulation where we assume a known thickness of upper layer but unknown thermal parameters. We used the FD forward model and optimized for diffusivity values - for the surface as well as the underneath layer, and the absorption coefficient of the surface layer. \cref{fig:2Layers}(c,d) show the recovered thermal properties for our simulations. We assumed a thickness of 0.5 mm and obtained the diffusivity values of the two layers with 0.01\% error. This shows promise in the approach and is a good direction for future work.

\section{Conclusion}
Tapping into the intrinsic properties of objects such as a material's thermodynamic constants ($k, \varepsilon$) enables robust classification, and alleviates some of the challenges faced by the more common visible camera-based approaches.
%
%
We proposed a simple optical setup consisting of a light source and thermal camera that captures the TSF of the scene.
We then used a novel FD-based iterative solver that robustly estimates $k,\varepsilon$ which was used for the downstream task of material classification.
Thermal material classification can be used as a complementary modality along with existing techniques including visible sensing. In biometrics, spoofing-based attacks such as using a mannequin for a real human become easier to detect due to the complementary strengths of thermal and visible cameras. 
Hence our work also has the potential to impact numerous areas in vision and security surveillance.

\vspace{15pt}
\noindent \textbf{Acknowledgements:} This work was supported by NSF awards IIS-2107313 and IIS-1730574, and ONR award N00014-21-1-2035. We thank Manuel Ballester for his help in formulating numerical methods for the heat equation, Rishubh Parihar for discussion on optimization, and Lionel Fiske for discussion on PDE solutions. We also thank the reviewers for their valuable suggestions to improve the quality of the manuscript.




{\small
\bibliographystyle{ieee_fullname}
\bibliography{_tsf.bib}

\begin{thebibliography}{10}\itemsep=-1pt

\bibitem{parida2012classification}
Patitapabana Parida, Ajit Behera, and S~Chandra Mishra.
\newblock Classification of biomaterials used in medicine.
\newblock {\em International Journal of Advances in Applied Sciences (IJAAS)},
  2012.

\bibitem{hendrix1996technologies}
Joerg Hendrix.
\newblock Technologies for the identification, separation and recycling of
  automotive plastics.
\newblock {\em Int. J. Environmentally Conscious Design \& Manufacturing},
  5(1):35--47, 1996.

\bibitem{buekens2014recycling}
Alfons Buekens and Xujian Zhou.
\newblock Recycling plastics from automotive shredder residues: A review.
\newblock {\em Journal of Material Cycles and Waste Management},
  16(3):398--414, 2014.

\bibitem{wang2009material}
Oliver Wang, Prabath Gunawardane, Steve Scher, and James Davis.
\newblock Material classification using brdf slices.
\newblock In {\em 2009 IEEE Conference on Computer Vision and Pattern
  Recognition}, pages 2805--2811. IEEE, 2009.

\bibitem{salamati2009material}
Neda Salamati, Cl{\'e}ment Fredembach, and Sabine S{\"u}sstrunk.
\newblock Material classification using color and nir images.
\newblock In {\em Color and Imaging Conference}, volume 2009, pages 216--222.
  Society for Imaging Science and Technology, 2009.

\bibitem{tanaka2017material}
Kenichiro Tanaka, Yasuhiro Mukaigawa, Takuya Funatomi, Hiroyuki Kubo, Yasuyuki
  Matsushita, and Yasushi Yagi.
\newblock Material classification using frequency-and depth-dependent
  time-of-flight distortion.
\newblock In {\em Proceedings of the IEEE Conference on Computer Vision and
  Pattern Recognition}, pages 79--88, 2017.

\bibitem{ibrahim2010spectral}
Abdelhameed~F Ibrahim, Shoji Tominaga, and Takahiko Horiuchi.
\newblock Spectral imaging method for material classification and inspection of
  printed circuit boards.
\newblock {\em Optical Engineering}, 49(5):057201, 2010.

\bibitem{saragadam2020programmable}
Vishwanath Saragadam and Aswin Sankaranarayanan.
\newblock Programmable spectrometry: Per-pixel material classification using
  learned spectral filters.
\newblock {\em IEEE Intl. Conf. Computational Photography (ICCP)}, 2020.

\bibitem{erickson2020multimodal}
Zackory Erickson, Eliot Xing, Bharat Srirangam, Sonia Chernova, and Charles~C
  Kemp.
\newblock Multimodal material classification for robots using spectroscopy and
  high resolution texture imaging.
\newblock In {\em 2020 IEEE/RSJ International Conference on Intelligent Robots
  and Systems (IROS)}, pages 10452--10459. IEEE, 2020.

\bibitem{saponaro2015material}
Philip Saponaro, Scott Sorensen, Abhishek Kolagunda, and Chandra Kambhamettu.
\newblock Material classification with thermal imagery.
\newblock In {\em Proceedings of the IEEE Conference on Computer Vision and
  Pattern Recognition}, pages 4649--4656, 2015.

\bibitem{bednarek2019robotic}
Jakub Bednarek, Michal Bednarek, Piotr Kicki, and Krzysztof Walas.
\newblock Robotic touch: Classification of materials for manipulation and
  walking.
\newblock In {\em 2019 2nd IEEE international conference on Soft Robotics
  (RoboSoft)}, pages 527--533. IEEE, 2019.

\bibitem{zhang2022thermal}
Jun Zhang, Han Li, Aiguo Song, Yizhuang Ding, and Juan Wu.
\newblock Thermal perception for information transmission: Theoretical
  analysis, device design, and experimental verification.
\newblock {\em IEEE Transactions on Haptics}, 2022.

\bibitem{guiatni2009thermal}
Mohamed Guiatni, Abdelaziz Benallegue, and Abderrahmane Kheddar.
\newblock Thermal display for telepresence based on neural identification and
  heat flux control.
\newblock {\em Presence: Teleoperators and Virtual Environments},
  18(2):156--169, 2009.

\bibitem{singhal2016development}
Anshul Singhal.
\newblock {\em Development of thermal displays for haptic interfaces}.
\newblock PhD thesis, Massachusetts Institute of Technology, 2016.

\bibitem{tilioua2018characterization}
Amine Tilioua, Laurent Libessart, and St{\'e}phane Lassue.
\newblock Characterization of the thermal properties of fibrous insulation
  materials made from recycled textile fibers for building applications:
  Theoretical and experimental analyses.
\newblock {\em Applied Thermal Engineering}, 142:56--67, 2018.

\bibitem{parker1961flash}
WJ Parker, RJ Jenkins, CP Butler, and GL Abbott.
\newblock Flash method of determining thermal diffusivity, heat capacity, and
  thermal conductivity.
\newblock {\em Journal of applied physics}, 32(9):1679--1684, 1961.

\bibitem{cowan1963pulse}
Robert~D Cowan.
\newblock Pulse method of measuring thermal diffusivity at high temperatures.
\newblock {\em Journal of Applied Physics}, 34(4):926--927, 1963.

\bibitem{garrido2020thermographic}
I Garrido, S Lag{\"u}ela, R Otero, and P Arias.
\newblock Thermographic methodologies used in infrastructure inspection: A
  review—post-processing procedures.
\newblock {\em Applied Energy}, 266:114857, 2020.

\bibitem{bergman2011introduction}
Theodore~L Bergman, Adrienne~S Lavine, Frank~P Incropera, and David~P DeWitt.
\newblock {\em Introduction to heat transfer}.
\newblock John Wiley \& Sons, 2011.

\bibitem{haberman2003applied}
Richard Haberman.
\newblock {\em Applied partial differential equations}, volume~4.
\newblock Prentice Hall Upper Saddle River, NJ:, 2003.

\bibitem{guo2017focal}
Qi Guo, Emma Alexander, and Todd Zickler.
\newblock Focal track: Depth and accommodation with oscillating lens
  deformation.
\newblock In {\em Proceedings of the IEEE international conference on computer
  vision}, pages 966--974, 2017.

\bibitem{alexander2016focal}
Emma Alexander, Qi Guo, Sanjeev Koppal, Steven Gortler, and Todd Zickler.
\newblock Focal flow: Measuring distance and velocity with defocus and
  differential motion.
\newblock In {\em European conference on computer vision}, pages 667--682.
  Springer, 2016.

\bibitem{morton2005numerical}
Keith~W Morton and David~Francis Mayers.
\newblock {\em Numerical solution of partial differential equations: an
  introduction}.
\newblock Cambridge university press, 2005.

\bibitem{Paszke_PyTorch_An_Imperative_2019}
Adam Paszke, Sam Gross, Francisco Massa, Adam Lerer, James Bradbury, Gregory
  Chanan, Trevor Killeen, Zeming Lin, Natalia Gimelshein, Luca Antiga, Alban
  Desmaison, Andreas Kopf, Edward Yang, Zachary DeVito, Martin Raison, Alykhan
  Tejani, Sasank Chilamkurthy, Benoit Steiner, Lu Fang, Junjie Bai, and Soumith
  Chintala.
\newblock {PyTorch: An Imperative Style, High-Performance Deep Learning
  Library}.
\newblock In H. Wallach, H. Larochelle, A. Beygelzimer, F. d'Alché Buc, E.
  Fox, and R. Garnett, editors, {\em Advances in Neural Information Processing
  Systems 32}, pages 8024--8035. Curran Associates, Inc., 2019.

\bibitem{landmann20183d}
Martin Landmann, Stefan Heist, Anika Brahm, Simon Schindwolf, Peter
  K{\"u}hmstedt, and Gunther Notni.
\newblock 3d shape measurement by thermal fringe projection: optimization of
  infrared (ir) projection parameters.
\newblock In {\em Dimensional Optical Metrology and Inspection for Practical
  Applications VII}, volume 10667, pages 9--18. SPIE, 2018.

\bibitem{landmann20203d}
Martin Landmann, Stefan Heist, Patrick Dietrich, Henri Speck, Peter
  K{\"u}hmstedt, Andreas T{\"u}nnermann, and Gunther Notni.
\newblock 3d shape measurement of objects with uncooperative surface by
  projection of aperiodic thermal patterns in simulation and experiment.
\newblock {\em Optical Engineering}, 59(9):094107, 2020.

\bibitem{erdozain20203d}
Jack Erdozain, Kazuto Ichimaru, Tomohiro Maeda, Hiroshi Kawasaki, Ramesh
  Raskar, and Achuta Kadambi.
\newblock 3d imaging for thermal cameras using structured light.
\newblock In {\em 2020 IEEE International Conference on Image Processing
  (ICIP)}, pages 2795--2799. IEEE, 2020.

\bibitem{raissi2019physics}
Maziar Raissi, Paris Perdikaris, and George~E Karniadakis.
\newblock Physics-informed neural networks: A deep learning framework for
  solving forward and inverse problems involving nonlinear partial differential
  equations.
\newblock {\em Journal of Computational physics}, 378:686--707, 2019.

\bibitem{cai2021physics}
Shengze Cai, Zhicheng Wang, Sifan Wang, Paris Perdikaris, and George~Em
  Karniadakis.
\newblock Physics-informed neural networks for heat transfer problems.
\newblock {\em Journal of Heat Transfer}, 143(6), 2021.

\bibitem{he2021physics}
Zhili He, Futao Ni, Weiguo Wang, and Jian Zhang.
\newblock A physics-informed deep learning method for solving direct and
  inverse heat conduction problems of materials.
\newblock {\em Materials Today Communications}, 28:102719, 2021.

\end{thebibliography}


\begin{thebibliography}{1}\itemsep=-1pt

\bibitem{smith2017cyclical}
Leslie~N Smith.
\newblock Cyclical learning rates for training neural networks.
\newblock In {\em 2017 IEEE winter conference on applications of computer
  vision (WACV)}, pages 464--472. IEEE, 2017.

\end{thebibliography}
}

\end{document}


\setcounter{figure}{9}

\definecolor{myorange}{RGB}{255, 165, 0}
\definecolor{mybrown}{RGB}{155, 78, 20}
\definecolor{mypurple}{RGB}{150, 56, 226}
\definecolor{myblue}{RGB}{10, 50, 230}
\definecolor{myred}{RGB}{230, 10, 10}
\definecolor{mygreen}{RGB}{60, 160, 60}
\definecolor{mydarkgreen}{RGB}{30, 110, 30}
\definecolor{myskyblue}{RGB}{31, 119, 180}
\definecolor{mymagenta}{RGB}{139, 0, 139}
\definecolor{mycyan}{RGB}{0, 100, 100}

\newcommand{\ad}[1]{{\color{mybrown}{[Aniket: #1]}}}
\newcommand{\vishwa}[1]{{\color{mygreen}{[Vishwa: #1]}}}
\newcommand{\emma}[1]{{\color{mypurple}{[Emma: #1]}}}
\newcommand{\florian}[1]{{\color{mymagenta}{[Florian: #1]}}}
\newcommand{\ashok}[1]{{\color{mydarkgreen}{[Ashok: #1]}}}
\newcommand{\aggelos}[1]{{\color{myblue}{[Aggelos: #1]}}}
\newcommand{\ollie}[1]{{\color{mypurple}{[Ollie: #1]}}}
\newcommand{\postrebtl}[1]{{\color{mydarkgreen}{[Post Rebuttal: #1]}}}
\newcommand{\delete}[1]{{\color{myred}{[DELETE: #1]}}}

\newcommand{\todo}[1]{{\color{myorange}{[TODO: #1]}}}
\newcommand{\help}[1]{{\color{myred}{[HELP: #1]}}}

\newcommand{\blue}[1]{{\color{myskyblue}{#1}}}
\newcommand{\orange}[1]{{\color{myorange}{#1}}}
\newcommand{\green}[1]{{\color{mygreen}{#1}}}

\def\epsP{\ensuremath{\epsilon^{\prime}}\xspace}
\def\fsx{{\ensuremath{f_s(x)}}\xspace}

\def\mtos{\ensuremath{\epsilon^{\prime}m^{2}/s}\xspace}


\title{Thermal Spread Functions (TSF): Physics-guided Material Classification \\
\large Supplementary - CVPR 2023}

\author{}

\maketitle
\section{Finite Difference Model - Performance}\label{sec:FDPerformance}

\begin{figure*}
\centering
\includegraphics[width=\textwidth]{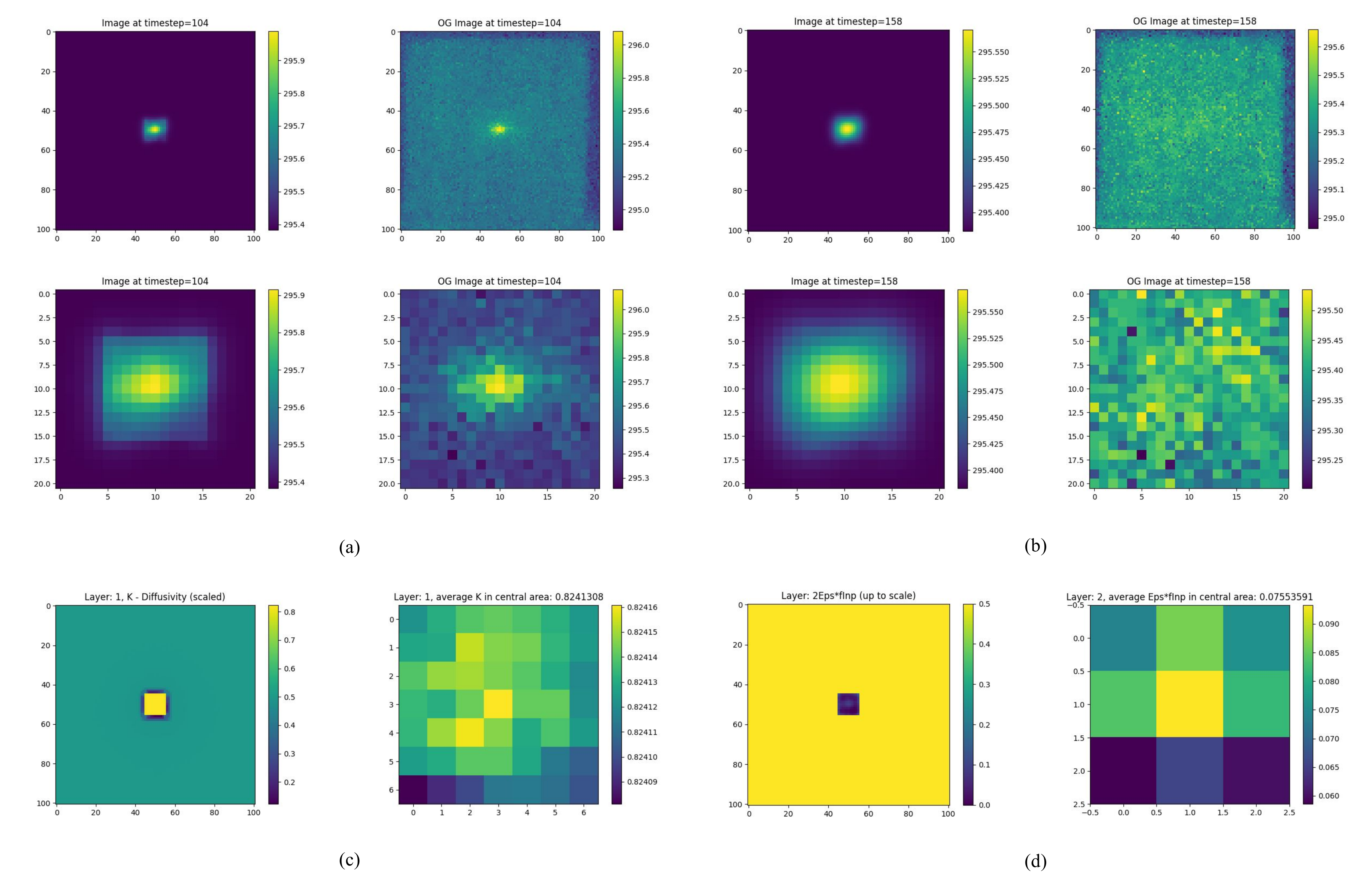}
\caption{\label{fig:FDPerformance}\textbf{Finite Difference Performance.} (a) \textbf{While the source is still on:} \textit{(Left)} Images obtained at time-step = 104, from FD method after optimization (Bottom row is the zoomed-in version of top row) \textit{(Right)} Original images captured from the thermal camera. (b) \textbf{After source is switched off:} Similar comparison of FD result and original captured image at time-step = 158. (c) The resultant diffusivity ($k$) image obtained after the optimization is complete, (left) original image (right) zoomed-in version (d) The resultant absorption coefficient (\epsP) image (left) original image (right) zoomed-in version. The results displayed are for Oakwood, average MSE error over the set of images is  8.68\time10 $^{-3}$ .}
\end{figure*}

\begin{figure}
    \centering
    \includegraphics[width=8cm]{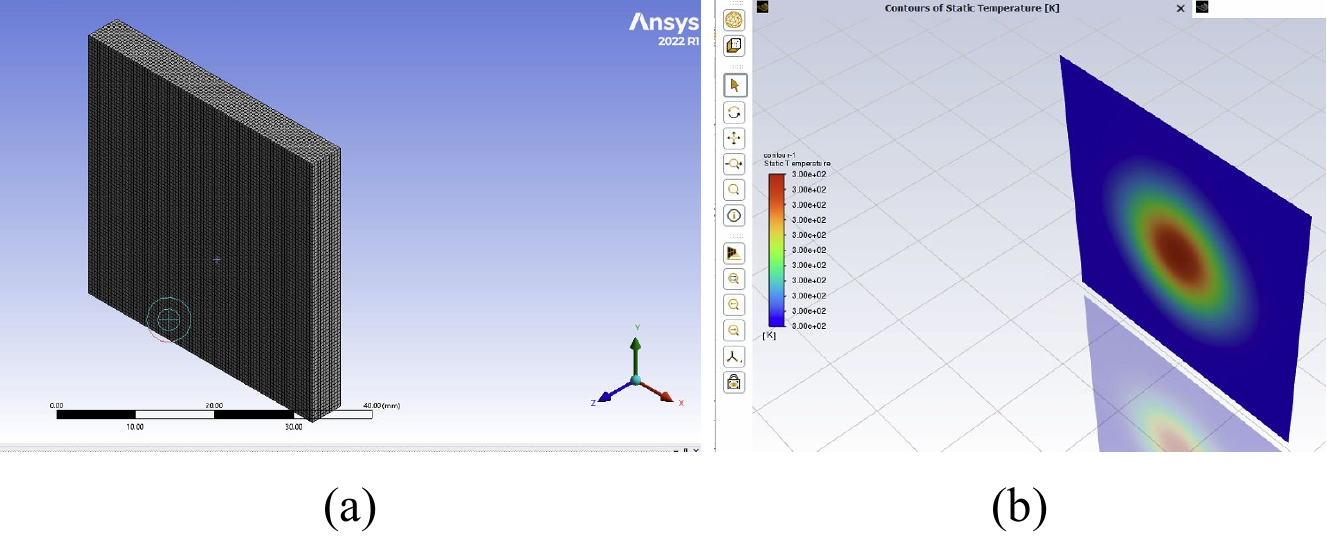}
    \caption{\textbf{Ansys Fluent FEM Analysis software.} (a) We give a custom heat injection on the surface of the material for few seconds and then turn it off. (b) We observe the surface temperature variation and do our analysis based on only surface temperatures which is what would be available to us during real measurements.}
    \label{fig:Simulations}
\end{figure}

\Cref{fig:FDPerformance} demonstrates the performance of our inverse Finite Difference algorithm for recovery of diffusivity $k$ and absorption coefficient \epsP. We show images at intermediate time-steps, one while the material gets heated (\Cref{fig:FDPerformance}(a)) and another image while it is cooling off after the source is turned off (\Cref{fig:FDPerformance}(b)). We also show our simulation setup in \cref{fig:Simulations}.

\section{Advantages over spectral methods}

For the PVC readings, we painted one of the faces with a highlighter (refer \Cref{fig:MetalsTaped}(a))  and tried our approach and it still worked quite well (refer to Fig. 7(b) of the main paper, where the classification accuracy is 100\% for PVC). This proves that small surface manipulations do not hinder our approach - unlike other spectral or RGB-image based methods.

\section{Tuning the hyper-parameter of learning rate}

Despite having automated everything else in our code, there is still one hyper-parameter that needs tuning in our setup - the learning rate for the optimization. We need to adjust the learning rate so that the optimization - (a) does not get stuck in a local minima and (b) does not wildly oscillate and hence not converge. Currently we do this manually by looking at the loss and diffusivity convergence curves but this process can be optimized as shown by previous work \cite{smith2017cyclical}

\section{Special case - metals}

We tried our approach on metals but it didn't work right off the bat because of two reasons - (1) The reflectivity of metals is too high and their emissivity is quite low. This means, it reflects most of the light we shine on it and whatever small fraction is absorbed and leads to a minor temperature change, is not visible because only a fraction of that reaches camera (because of low emissivity) (2) The diffusivity of metals is very high. Combined with it's high reflectivity and low emissivity, we need a much higher power laser compared to the one we use currently (60mW).

\begin{figure}
\centering
\includegraphics[width=0.5\textwidth]{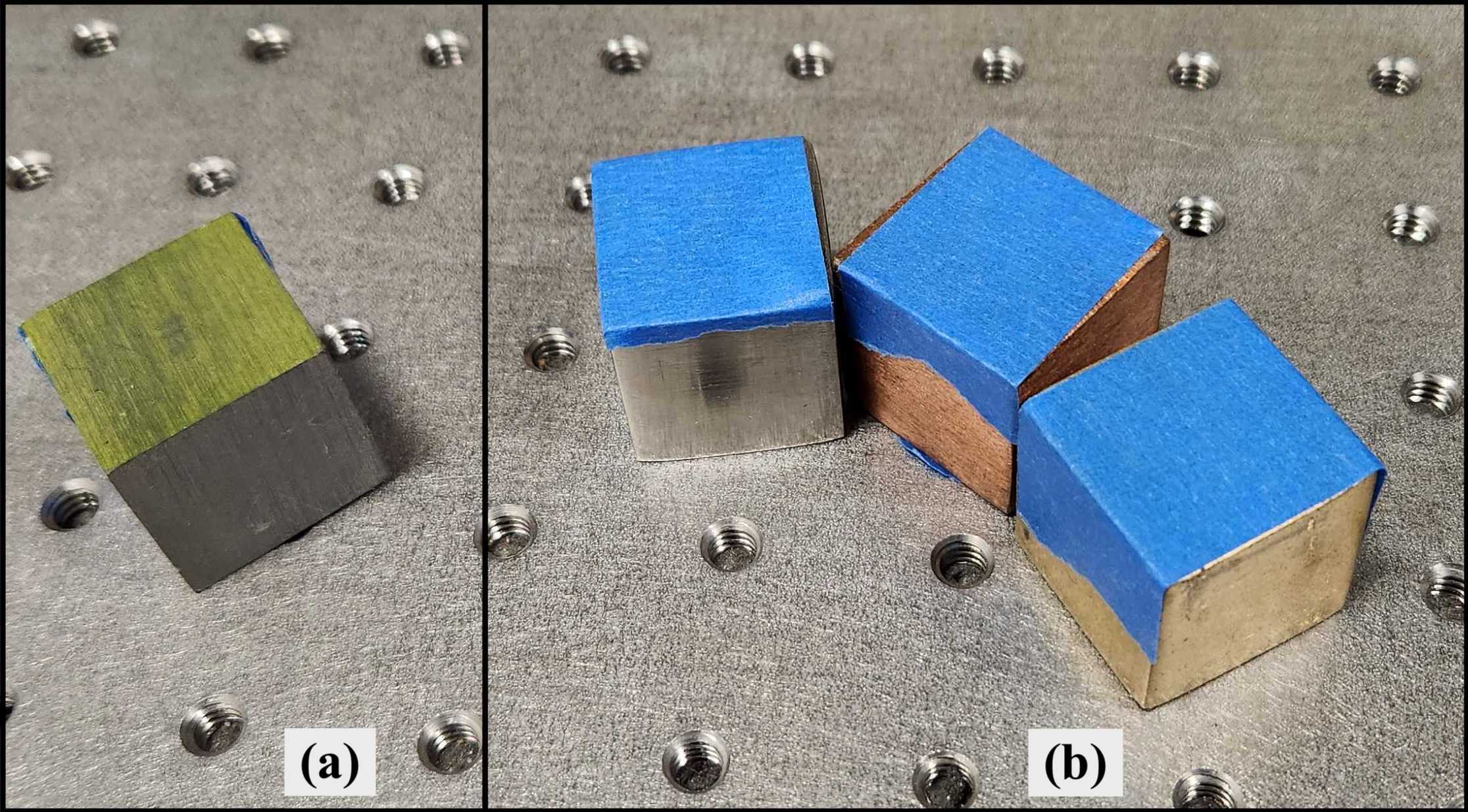}
\caption{\label{fig:MetalsTaped}\textbf{Metals Taped.} (a) We painted a PVC block with a green highlighter and tried the same approach and our algorithm correctly classifies the material. (b) We pasted a Scotch blue paper-tape on top of the metals - (left-to-right) Stainless Steel, Copper and Brass.}
\end{figure}

\begin{figure*}
    \centering
    \includegraphics[width=\linewidth]{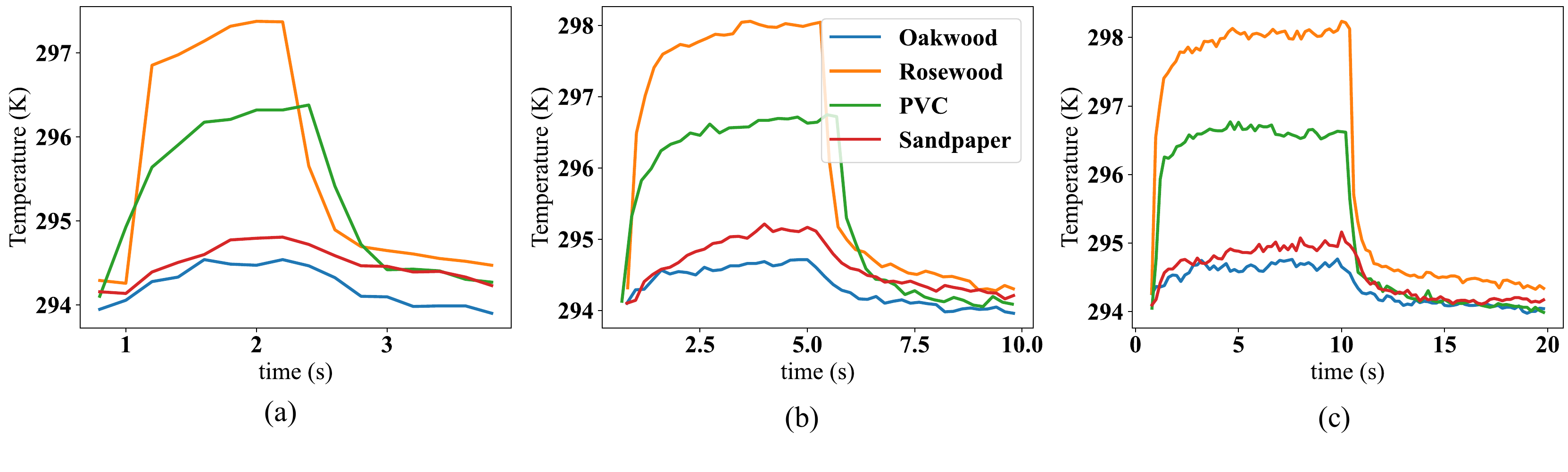}
    \caption{\textbf{TSFs for various scan durations.} TSFs plotted for Oakwood, Rosewood, PVC, and Sandpaper for time durations (a) 4s, (b) 10s, (c) 20s}
    \label{fig:MultipleDurations}
\end{figure*}

\begin{figure}
\centering
\includegraphics[width=0.5\textwidth]{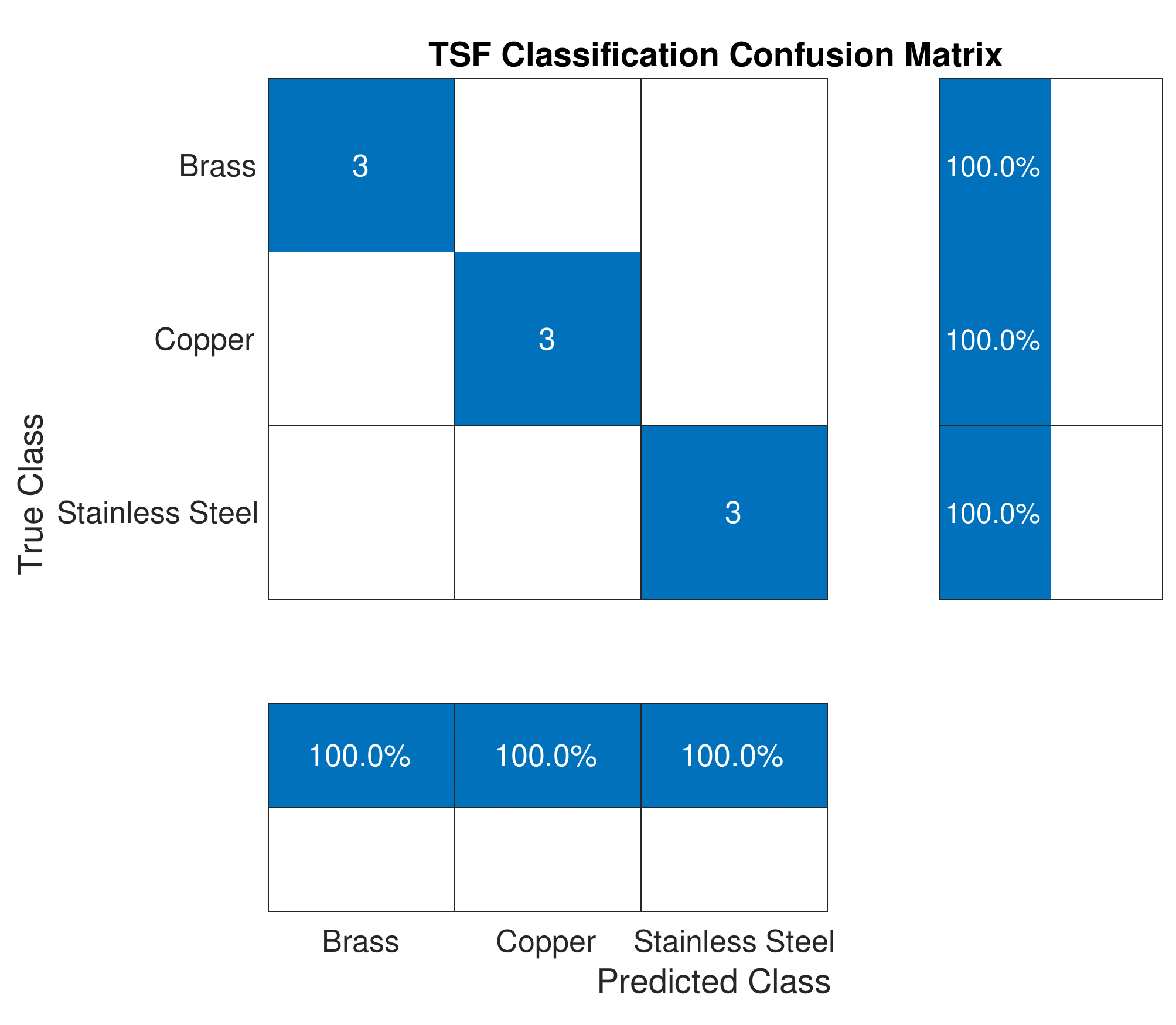}
\caption{\label{fig:CM_metals}\textbf{Confusion Matrix for metals.} We perform leave-one-out validation to gauge the accuracy of our classification using TSF.}
\end{figure}

\textbf{Workaround}: We tested another method to still get some classification means for metals for the low powered laser. We pasted a tape on top of the metal cubes. This gives us the advantage of low reflectivity and higher emissivity. Thus, we might get absorption properties of the tape but the diffusivity for the metal can be obtained in this method. Please find the setup in \Cref{fig:MetalsTaped}(b). The confusion matrix obtained upon classification for these metal cubes is shown in \Cref{fig:CM_metals}.

\section{Two layers - going under the surface}

\begin{figure}
\centering
\includegraphics[width=0.5\textwidth]{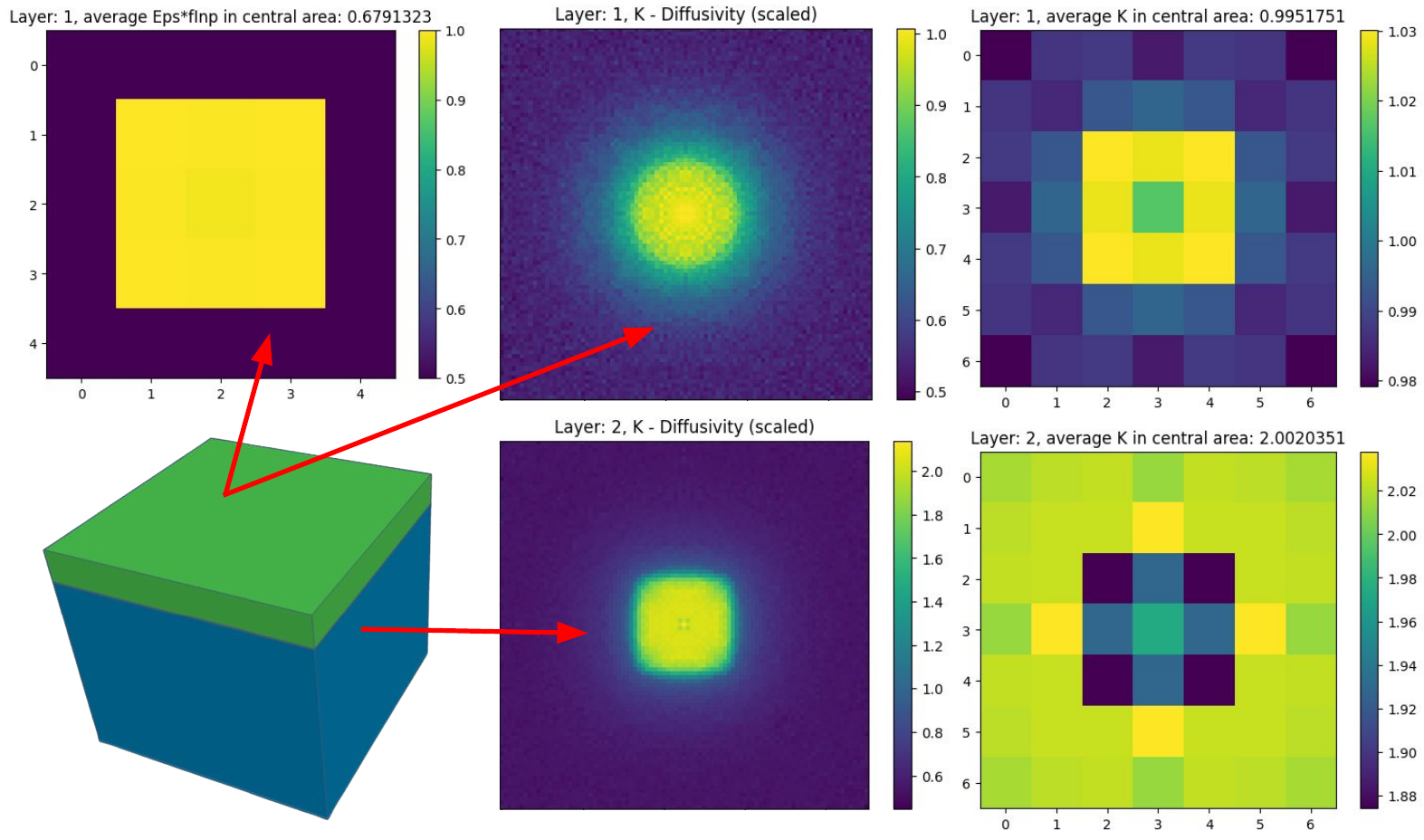}
\caption{\label{fig:TwoLayers}\textbf{Recovering properties of hidden material.} (Top) Recovered properties of the top layer - absorption coefficient \epsP, thermal diffusivity $k$ and its zoomed-in version. (Bottom) Two-layered material, and properties of the bottom layer - thermal diffusivity $k$ and its zoomed-in version.}
\end{figure}

We performed simulations to test out our hypothesis of using this approach to find properties of under-the-surface materials. We created a heat diffusion simulation using our forward FD model, where the surface layer with a thickness of 0.5mm, has a diffusivity of 1\time10$^{-7}$ \mtos. The layer below it has a diffusivity of 2\time10$^{-7}$ \mtos. We assume the thickness of top layer as a known parameter for this optimization. After running our optimization algorithm for recovering 2 layer properties, we got a diffusivity for top layer as 0.995\time10$^{-7}$ \mtos and a value of 2.002\time10$^{-7}$ \mtos for the bottom layer (refer \Cref{fig:TwoLayers}). These values match very closely with the original values which makes it a good direction to work on in the future.

\section{Varying capture time $t_{ON}$}

Apart from the results shown in the paper, we also tried various smaller capture times for our setup. The TSFs are similar for the materials for all the $t_{ON}$'s (refer \cref{fig:MultipleDurations}). We found that our approach resulted in similar results for a total capture duration of 20s or higher. While this duration can be further reduced with implicit approaches such as PINN, the physical process of heat dissipation requires a minimum duration, which often tends to be several tens of seconds. 

{\small
\bibliographystyle{ieee_fullname}
\bibliography{refs.bib}
}